%% file: nips_2018.tex
\documentclass{article}

\usepackage[square, numbers]{natbib}

\usepackage[preprint]{nips_2018}

\usepackage[utf8]{inputenc} 
\usepackage[T1]{fontenc}    
\usepackage[russian, english]{babel}
\usepackage{hyperref}       
\usepackage{url}            
\usepackage{booktabs}       
\usepackage{amsfonts}       
\usepackage{nicefrac}       
\usepackage{microtype}      
\usepackage{amsmath} 
\usepackage{graphicx} 
\usepackage{algorithm}  
\usepackage{algorithmic}  
\usepackage{mathtools} 
\usepackage{color}
\usepackage[title]{appendix}
\usepackage{longtable}
\usepackage{marginnote}
\usepackage{tempora}

\newcommand{\x}{{\mathbf x}}
\newcommand{\y}{{\mathbf y}}
\newcommand{\ru}{\selectlanguage{russian}}
\newcommand{\cmmnt}[1]{}

\title{Zero-Shot Dual Machine Translation}

\author{
  \And
  Lierni Sestorain \\
  ETH Zürich \\
  \texttt{liernis@student.ethz.ch} \\
  \And
  Massimiliano Ciaramita \\
  Google \\
  \texttt{massi@google.com} \\
  \And
  Christian Buck \\
  Google \\
  \texttt{cbuck@google.com} \\
  \And
  Thomas Hofmann \\
  ETH Z\"urich \\
  \texttt{thomas.hofmann@inf.ethz.ch} \\ 
}

\begin{document}

\maketitle

\input{abstract}
\input{introduction}
\input{relatedWork}

\input{model}

\input{experiments}
\input{conclusion}


\bibliographystyle{abbrvnat}
\bibliography{refs}

\begin{appendices}
\input{qualitative}

\end{appendices}

\end{document}

%% file: abstract.tex

\begin{abstract}
Neural Machine Translation (NMT) systems rely on large amounts of parallel data. 
This is a major challenge for low-resource languages. 
Building on recent work on unsupervised and
semi-supervised methods, we present an approach that combines zero-shot and dual learning.
The latter relies on reinforcement learning, to exploit the duality of the machine translation task, and requires only monolingual data for the target language pair.
Experiments show that a zero-shot dual system, trained on
English-French and English-Spanish, outperforms by large margins 
a standard NMT system in zero-shot translation performance on Spanish-French
(both directions).
The zero-shot dual method approaches the performance, within
2.2 BLEU points, of a comparable supervised setting.
Our method can obtain improvements also on the setting where a small amount of parallel data for the zero-shot language pair is available.
Adding Russian, to extend our experiments to jointly modeling $6$
zero-shot translation directions, all directions improve between $4$ and
$15$ BLEU points, again, reaching performance near that of the supervised setting.
\end{abstract}

%% file: introduction.tex

\section{Introduction}

The availability of large high-quality parallel corpora is key to building robust machine translation systems.
Obtaining such training data is often expensive or outright infeasible, thereby constraining the availability of translation systems, especially for low-resource language pairs.
Among the work trying to alleviate this data-bottleneck problem, \citet{johnson2016google} present a system that performs \emph{zero-shot} translation.
Their approach is based on a multi-task model trained to translate 
between multiple language pairs with a single encoder-decoder 
architecture. Multi-tasking is enabled simply by a special symbol,
introduced at the beginning of the source sentence,
that identifies the desired target language. 
Their idea enables a NMT model~\cite{wu2016google} to 
translate between language pairs never encountered during training, 
as long as source and target language were part of the training data.

Since any machine translation task has a canonical
inverse problem, that of translating in the opposite direction, 
\citet{he2016dual} propose a dual learning method for NMT. 
They report improved translation quality of two dual, weakly trained
translation models using \emph{monolingual} data from both languages.
For that, a monolingual sentence is translated twice, first into the target 
language and then back into the source language. 
A language model is used to measure the \emph{fluency} of the intermediate translation. 
In addition, a \emph{reconstruction} error is computed via back-translation. 
Fluency and reconstruction scores are combined in a reward function 
which is optimized with reinforcement learning.

We propose an approach that builds upon the multilingual NMT 
architecture~\citep{johnson2016google} to improve the zero-shot 
translation quality by applying the reinforcement learning ideas 
from the dual learning work~\citep{he2016dual}. 
Our model outperforms the zero-shot system's performance 
while being simpler and more scalable than the original dual formulation,
as it learns a single model for all involved translation directions. 
Moreover, for the original dual method to work, some amount of parallel data 
is needed in order to train an initial weak translation model.
If no parallel data is available for that language pair, 
the original dual approach is not feasible,
as a random translation model is ineffective 
in guiding exploration for reinforcement learning.
In our formulation, however, no parallel data is required at all
to make the algorithm work, as the zero-shot mechanism kick-starts the dual learning. 

Experiments show that our approach outperforms the zero-shot translation 
of the NMT model by up to $32.58$ BLEU points using only monolingual corpora. 
Furthermore, our system leads to gains of 0.67-2.53 BLEU points
when the base NMT model benefits from a small parallel data 
for the zero-shot translation pair.
We also investigate a setting where six new language pairs are learned, 
by adding Russian to the set of languages. 
These experiments produce similar improvements, 
and also shed additional light on the zero-shot learning process.

%% file: relatedWork.tex

\section{Related work}
\label{ch:sota}
We build on the standard encoder-attention-decoder architecture for
neural machine
translation~\citep{sutskever2014sequence,cho2014learning,bahdanau2015},
and specifically on the Google NMT implementation~\citep{wu2016google}.
This architecture has been extended by
\citet{johnson2016google} for unsupervised or zero-shot
translation.
Here, one single multi-task system is 
trained on a subset
of language pairs, but then can be used to translate between all
pairs, including unseen ones. The success of this approach will depend
on the ability to learn interlingua semantic representations of which
some evidence is provided in \citep{johnson2016google}. In a zero-shot evaluation from Portuguese to Spanish \citet{johnson2016google} report a gap of 6.75 BLEU points from the supervised NMT model.

\citet{he2016dual} propose a learning
mechanism that bootstraps an initial low-quality translation model.
This so-called dual learning approach leverages criteria to
measure the fluency of the translation and the reconstruction error of
back-translating into the source language.
As shown in \citep{he2016dual}, this can be effective,
provided that the initial translation model provides a sufficiently good starting point.
In the reported experiments, they bootstrap an NMT model~\citep{bahdanau2015} trained on $1.2$M parallel sentences. Dual learning improves the BLEU score on WMT'14 French-to-English by $5.23$ points, obtaining comparable accuracy to a model trained on 10x more data, i.e.~$12$M parallel sentences.

\citet{xia2017dual} use dual learning in a supervised setting and
improve WMT'14 English-to-French translation by $2.07$ BLEU points over the baseline of \cite{bahdanau2015,jean2014using}. 
 In the supervised setting~\citet{Bahdanau:2017} show that Actor-Critic
can be more effective than vanilla policy gradient~\citep{Williams:1992}.

Recent work has also explored fully unsupervised machine
translation. 
\citet{lample2017unsupervised} propose a model that takes
sentences from monolingual corpora from both languages and maps them
into the same latent space. 
Starting from an unsupervised word-by-word
translation, they iteratively train encoder-decoder to reconstruct and translate from a noisy version of the input.
Latent distributions from the two domains are aligned using adversarial
discriminators. 
At test time, encoder and decoder work as a standard NMT
system. 
\citet{lample2017unsupervised} obtain BLEU score of $15.05$ on WMT'14 English-to-French translation, improving over a word-by-word translation baseline with inferred bilingual dictionary \cite{conneau2017word} by $8.77$ BLEU points.

\citet{artetxe2017unsupervised} present a method related to~\citep{lample2017unsupervised}.
The model is an encoder-decoder system with attention trained to perform
autoencoding and on-the-fly back-translation. 
The system handles both translation directions at the same time. 
Moreover, it shares the encoder across languages, in order to
generate language independent representations, by exploiting pre-trained cross-lingual embeddings. 
This approach performs comparably on WMT'14 English-to-French, improving by $0.08$ BLEU points over \cite{lample2017unsupervised}. Furthermore, WMT'14 French-to-Spanish translation obtains $15.56$ BLEU points.

\citet{lample2018phrase} combine the previous two approaches. 
They first learn NMT language models, as denoising autoencoders, 
over the source and target languages.
Leveraging these, two translation models are initialized, 
one in each direction.
Iteratively, source and target sentences are translated using the current
translation models and 
new translation models are trained on the generated pseudo-parallel sentences.
The encoder-decoder parameters are shared across the two languages
to share the latent representations and for regularization, respectively.
Their results outperform those of \cite{lample2017unsupervised,artetxe2017unsupervised}, 
obtaining $25.1$ BLEU points on WMT'14 English-to-French translation. BLEU scores are still distinctly lower than the corresponding supervised model.\footnote{Exact scores are not reported for the supervised comparison, but see Figure~2 in~\citep{lample2018phrase}.}

%% file: model.tex

\section{Zero-shot dual machine translation}
\label{ch:model}

Our method for unsupervised machine translation works as follows:
We start with a multi-language NMT model as used in zero-shot translation \citep{johnson2016google}. 
However, we then train this model using
only monolingual data in a dual learning framework similar to that proposed
by~\citet{he2016dual}.

At least three languages are involved in our setting, later we will also
experiment with four. Let them be X, Y
and Z. Let the pairs Z-X and Z-Y be the language pairs with available
parallel data and X-Y be the target language pair without. 
We thus assume to have access to sufficiently large parallel corpora
$D_{Z\leftrightarrow X}$ and $D_{Z\leftrightarrow Y}$, 
monolingual data $D_X$ and
$D_Y$ for training the language models, 
as well as a small corpus $D_{X\leftrightarrow Y}$
for the low-resource pair evaluation. 

\subsection{Zero-shot dual training}
A single multilingual NMT model with parameters $\theta$ is trained
on both $D_{Z\leftrightarrow X}$ and $D_{Z\leftrightarrow Y}$, 
in both directions. 
We stress that this model does not train on any X-Y sentence
pairs. The multilingual NMT model is capable of generalizing to
unseen pairs, that is, it can translate from X to Y and vice
versa. Yet, the quality of the zero-shot pt$\rightarrow$es translations, in the best setting (Model 2 in~\citep{johnson2016google}) is approximately 7 BLEU points lower than the corresponding supervised model.

Using the monolingual data $D_X$ and $D_Y$, we train two disjoint
language models with maximum likelihood.
We implement the language models as multi-layered LSTMs\footnote{\url{https://www.tensorflow.org/tutorials/recurrent}} that process one word at a time. 
For a given sentence, the language model outputs a probability $P_X(.)$, 
quantifying the degree of fluency of the
sentence in the corresponding language X.
Similar to \citep{he2016dual}, we define a
REINFORCE-like learning algorithm~\citep{Williams:1992}, where the
translation model is the \emph{policy} that we want to learn. 
The procedure on one sample is, schematically:
\begin{enumerate}
\item Given a sentence $\x$ in language X, sample a corresponding
  translation from the multilingual NMT model $\y \sim P_{\theta}(\cdot|\x)$.
\item Compute the fluency reward, $r_1 = 
\log P_Y(\y)$. 
\item Compute the reconstruction reward $r_2 = \log P_{\theta}(\x | \y)$ by using $P_{\theta}(\cdot |{\mathbf y})$ for back-translation.
\item Compute the total reward for $\y$ as $R= \alpha r_1 + (1 - \alpha ) r_2 $, where $\alpha$ is a hyper-parameter.
\item Update $\theta$ by scaling the gradient of the cross entropy
loss by the reward $R$.
\end{enumerate}

The process can be started with a sentence from either language X or Y and works symmetrically.

\subsection{Details of the gradient computation}
The reward for
generating the target $\y$ from source $\x$ is defined as:

\begin{equation}
R(\y) = \alpha \log P_Y(\y) + (1 - \alpha) \log P_{\theta}(\x | \y)
\end{equation}

We want to optimize the expected reward, under the translation model
(the policy) $P_\theta$ and thus compute the
gradient with respect to the expected reward:

\begin{align}
\nabla_{\theta} \mathbb{E}_{\y|\x}[R] 
& = \nabla_{\theta} \sum_{\y}  P_\theta(\y | \x) R(\y) 
= \sum_{\y} \nabla_{\theta}  P_{\theta}(\y | \x) R(\y) + P_{\theta}(\y | \x) \nabla_{\theta} R(\y) \\
&= \mathbb{E}_{\y|\x} \left[  R(\y) \nabla_{\theta} \log P_{\theta}(\y | \x)    +    (1 - \alpha) \nabla_{\theta}    \log P_{\theta}(\x | \y)  \right]
\end{align}
Here we have used the product rule and the identity $\nabla P = P \cdot \nabla \log P$. Notice that, the reward depends (differentiably) on $\theta$, because of the reconstruction term.

Since the gradient is a conditional expectation, one can sample $\y$ to obtain an unbiased estimate of the expected gradient~\citep{sutton2000policy}:
\begin{equation}
\nabla_{\theta} \mathbb{E}_{\y|\x}[R]  \approx \frac{1}{K} \displaystyle \sum_{i} 
R(\y_i)  \nabla_{\theta} \log P_{\theta}(\y_i | \x)   + (1 - \alpha) 
  \nabla_{\theta} \log P_{\theta}(\x | \y_i)
\end{equation}

Any sampling approach can be used to estimate the expected gradient. 
\citet{he2016dual} use beam search instead of sampling to avoid large variance and unreasonable results. 
We generate from the policy by randomly sampling the next word from the predicted word distribution.
However, we found it helpful to make the distribution more deterministic using a low softmax temperature ($0.002$). 
 In order to reduce variance further, we compute a baseline through the batch. 
That is, we compute the average of each reward through the whole batch and subtract it from each reward.
We optimize the likelihood of the translation system on the monolingual data of both the source and target languages, while scaling the gradients (equivalently, the loss or the negative likelihood) with the total rewards.

\subsection{System implementation}

\begin{figure}
\centering
\includegraphics[scale=0.67]{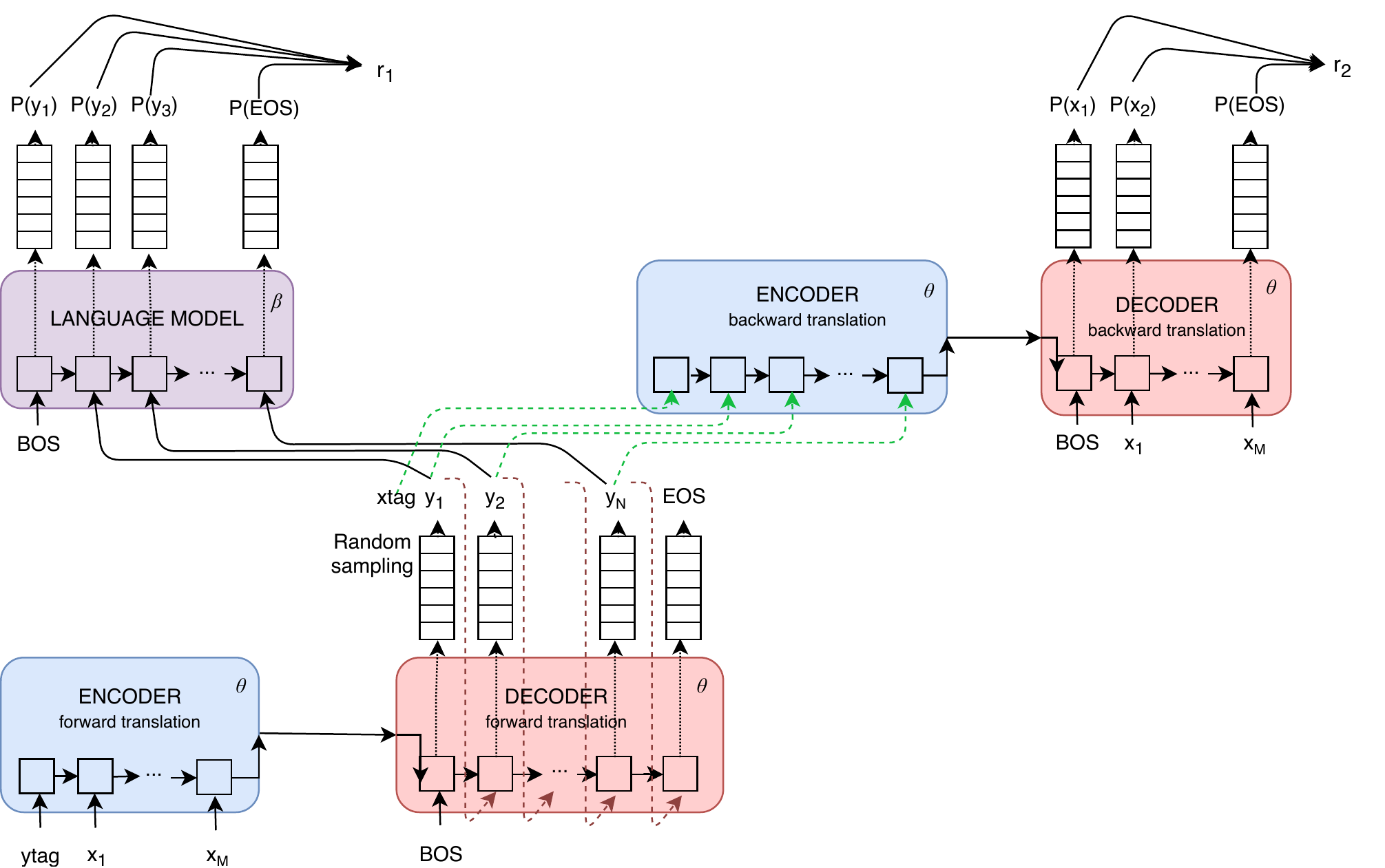}
\caption{The zero-shot dual learning procedure (bottom-up,
  left-to-right). The encoder, on the 
  forward translation step, receives the source sentence $\x$
  as input with the target language tag $y_{tag}$. The decoder, on the
  forward translation step, randomly samples a translation
  $\y$, which is passed to the language model to
  generate the fluency reward. At the same time, $\y$
  is encoded, by the same model, now with the source language tag
  $x_{tag}$. Then the decoder is used to compute the probability of
  the original sentence $\x$, conditioning on
  $\y$. Both encoders and decoders share the same parameters.} 
\label{fig:model_arch}
\end{figure}

The system is implemented as an extension of the Tensorflow \cite{abadi2016tensorflow} NMT tutorial
code,\footnote{\url{https://www.tensorflow.org/tutorials/seq2seq}.} and will be released as open source.
Figure~\ref{fig:model_arch} visualizes the training
process. 
The model is optimized after two calls: one for the forward translation and another for the backward translation.
During the forward translation, the encoder takes a sentence $\x=y_{\mathrm{tag}}, x_1, ..., x_M$, where $y_{\mathrm{tag}}$ is the tag of the
target language, and
samples a translation output $\y = y_1, y_2, ..., y_N$.  At each step the decoder randomly picks the next word from the current distribution.  This word is fed as input at the next decoding step. 

Then, $\y$ is used by the language model RNN trained on the target language. 
The language model is kept fixed during the whole process. 
Observe that initially the special begin-of-sentence symbol is fed, 
so that we can also obtain the probability for the first word. 
Here, we use the output logits in order to calculate the probability of the sample sentence. The first reward is computed as:

\begin{equation}
  r_1 = \log P_{Y}(\y) = \displaystyle \sum_{i=1}^N \log  P_{Y}(y_i | y_1, \ldots, y_{i-1}) 
\end{equation}

In addition, the backward translation call is executed: 
 $\y$, plus the source language tag added 
in the beginning, $ x_{\mathrm{tag}} $, is encoded to a sequence of hidden state
vectors 
$h_0, h_1, h_2, \ldots, h_N$
using the encoder.
These outputs are passed to the decoder which is fed with the original source words $x_i$ at each step. 
The output logits are used to calculate the probability of the source 
sentence $\x$ being reconstructed from $\y$. 
This value is used as the second reward:

\begin{equation}  
r_2 = \log P_{\theta}(\x| \y) = \sum_{i=1}^M \log P_{\theta}(x_i | x_1, \ldots, x_{i-1}, h_0,
h_1, \ldots, h_N)
\end{equation}

%% file: experiments.tex

\section{Experiments}
\label{ch:experiments}

\subsection{Dataset}
A key factor in the choice of dataset was the fact that we require parallel data for the supervised language pairs (Z-X, Z-Y),
in order to perform zero-shot translation (X-Y). 
This limited us from working with the widely used WMT dataset. 
We chose the United Nations Parallel Corpus (UN corpus) \cite{ziemski2016united}, instead.
The corpus is composed of official records and other parliamentary documents of the United Nations for the six official UN languages:
Arabic (ar), Chinese (zh), English (en), French (fr), Russian (ru) and Spanish (es). 
The corpus contains pairwise aligned documents as well as a fully aligned sub-corpora for the six languages, 
thus it allows the control needed for our experiments,
without having to resort on human ratings.
Moreover, the corpus provides official development and test sets composed of the documents released in 2015. 
Both sets comprise 4,000 sentences, aligned for all the six languages. 
This allows experiments to be evaluated, and replicated, in all directions.

\subsection{Vocabulary}

As in \cite{johnson2016google}, we segment the data using an algorithm similar 
to the Byte-Pair-Encoding (BPE) \cite{sennrich2015neural} in order to efficiently deal with unknown words while still keeping a limited vocabulary. 
This approach offers a good balance between the flexibility of character-delimited models and the efficiency of word-delimited models. 
The corpus is initially represented using a character vocabulary. 
The vocabulary is iteratively updated by replacing the most frequent symbol pair on the corpus with a new symbol,
which adds one new element to the vocabulary.

We calculate the vocabulary on the datasets sampled for the translation system.
We apply 32K merge operations which yields a vocabulary of roughly 33K sub-word units, shared across the three languages. 
Every model that we train --the language models and the translation systems-- 
relies on this vocabulary, unless explicitly stated. 
Three language tags are added at the end of the vocabulary in the translation systems for the zero-shot purpose 
(the language model does not use them).

\subsection{Language models}
We picked 4 million random sentences from the UN corpus for each
language to train the language models.  
We used the training datasets of each language for that. In other
words, we picked the 4 million Spanish sentences from the union of the
Spanish files contained in the English-Spanish and Spanish-French
corpora. We picked the French training dataset symmetrically from the
English-French and Spanish-French files. 
We remove sentences longer than 100 words.

The language model is a 2-layer, 512-unit LSTM, with 65\% dropout probability applied on all non-recurrent connections.
It is trained using Stochastic Gradient Descent (SGD) to minimize the negative log probability of the target words.
We trained the model for six epochs, halving the learning rate, initially $1.0$, in every epoch starting at the third. The norm of the gradients is clipped at 10.
We used a batch size of 64 and the RNN is unrolled for 35 time steps.
The vocabulary is embedded to 512 dimensions.
Each language model was trained on a single Tesla-K80 for approximately 60 hours. 

Our Spanish language model obtains a wordpiece-level perplexity of $18.919$, while the French one obtains $15.090$. 
Since the perplexities we report are at sub-word unit level, it is not easy to quantify their quality. For that reason, we use the KenLM implementation \cite{heafield2011kenlm} with the same sample and  vocabulary. We obtain test perplexities of $14.76$ and $12.84$ for the Spanish and French language models, respectively. Even though the difference is significant, we do not focus on improving the language models as they are sufficient to obtain a coarse fluency metric. 

\subsection{Translation systems}
For the translation systems, we randomly sampled $1$M parallel
sentences for each language pair.  All sentences are disjoint across
language pairs. We removed
sentences longer than 100 tokens. Development and test sets are used
as they are in both cases. 

For our base model, we used a NMT with LSTM cells of 1024 hidden units and 8 layers as described in \cite{wu2016google}.
We have re-used their choice of hyper-parameters\footnote{\url{https://github.com/tensorflow/nmt/blob/master/nmt/standard_hparams/wmt16_gnmt_8_layer.json}}, except for the learning rate, which is initialized at $1.0$ and halved every $17$K steps. 
The baseline zero-shot NMT was trained in a single Tesla-P100 for approximately one 
week. 

Training with reinforcement learning is performed on 1M
\emph{monolingual} sentences, 500k for each Spanish and
French. This data does not include aligned pairs across languages. They are also disjoint from the ones used to train the base NMT model.
The model configuration is the same as the NMT base model, 
except for the batch size which needs to be halved for memory resources, given that we now need to load two language models.
We use learning rates 
$\gamma_{1,t}= 0.0002$ and $\gamma_{2,t} = 0.02$ and $\alpha$=0.005,  
as in \cite{he2016dual}. We did not attempt to optimize hyper-parameters to fine tune the model.
Training takes approximately 3 days on a single
Tesla-P100.\footnote{All experiments were carried out on the Google
  Cloud Platform.}

\section{Results}

\hspace{-0.3cm}
\begin{table}
\centering
  \begin{tabular}{lcccccc}
    \toprule
          & \bf Phrase-based   & \bf NMT-0 & \bf NMT-S & \bf NMT-F & \bf Dual-0  & \bf Dual-S  \\
         \midrule
    Aligned & en-fr (11M)& en-fr (1M) & en-fr (1M)  & en-fr (1M)  & en-fr (1M) & en-fr (1M)  \\
    
    Data&  en-es (11M)& en-es (1M)  & en-es (1M) & en-es (1M) & en-es (1M) & en-es (1M)   \\
     & es-fr (11M) & & es-fr (10k) & es-fr (1M) & & es-fr (10k) \\
    \midrule
    Monol.& & & & & es (0.5M) & es (0.5M) \\
    Data& & & & & fr (0.5M) & fr (0.5M) \\
    \midrule
    en$\rightarrow$es & 61.26       & 49.00 & 43.33   & 44.06   &
    37.05 \cmmnt{36.42}       & 38.74 \cmmnt{36.29} \\
    es$\rightarrow$en & 59.89       & 49.67 & 40.17   & 18.24   &
    32.84 \cmmnt{30.82}   & 32.03 \cmmnt{27.16}    \\ 
    en$\rightarrow$fr & 50.09       & 37.88 & 33.71   & 34.75   &
    29.58 \cmmnt{29.21} & 30.89 \cmmnt{29.22}      \\ 
    fr$\rightarrow$en & 52.22       & 42.12 & 34.17   & 13.58   &
    27.95 \cmmnt{25.52}   & 26.00 \cmmnt{19.96}    \\ 
    \midrule
    \midrule
    es$\rightarrow$fr & 52.44       & 10.02 & 33.10  & 37.67    
    & 35.54 \cmmnt{35.83} & 35.63 \cmmnt{36.47}     \\ 
    fr$\rightarrow$es & 49.79       &  6.25 & 38.33  & 40.85    &
    38.83 \cmmnt{38.91}  & 39.00 \cmmnt{39.72}      \\ 
    \bottomrule
  \end{tabular}
    \caption{BLEU scores on the UN corpus test set. Each line reports
      the BLEU scores of the corresponding translation
      direction. The first column refers to the phrase-based model
      of~\citet{ziemski2016united}.
      All NMT and Dual models
      are trained on 1M en-es and 1M en-fr aligned sentences, used in
      both directions. The NMT-S (small) model is trained additionally on 10k
      es-fr aligned sentences, while NMT-F (full) is trained additionally on
      1M es-fr sentences. The Dual-0 model does not use any es-fr
      aligned data, while Dual-S is trained starting from NMT-S.}   
  \label{tab:results-3lang}
\end{table}

\subsection{Supervised performance}
We report the BLEU score on the official test sets, 
as computed by \texttt{multi-bleu-detok.perl} script, 
downloaded from the public implementation of Moses\footnote{\url{http://www.statmt.org/moses/}}.
The first column of Table \ref{tab:results-3lang} reports the BLEU 
scores from~\citep{ziemski2016united} for a phrase-based translation system.
Notice that these models are trained on the complete fully-aligned UN data, 
that is, 11M sentence pairs for each language pair.
The second column reports the scores of our multilingual NMT baseline model (NMT-0). 
This is trained on 1M en-es and 1M en-fr parallel sentences. It is not trained on any es-fr data.

The first four lines of BLEU scores (en-es, en-fr) show that, in the supervised setting, the phrase-based model outperforms NMT-0 by more than 10 points.
The amount of training data obviously affects the model accuracy, but
we speculate that lower performance, with respect to the phrase-based systems, is also due to model capacity. 
All NMT models learn to translate in six different directions at the same time 
-- four in the case of NMT-0.   
The number of parameters per language/direction is decreased as the number of languages increases. 
In order to check this hypothesis, we have performed the following experiment. 
We trained a model with the same parameters for only one translation direction: en$\rightarrow$es. 
We built a specific vocabulary only on the data of this language pair. 
In this case, we obtain a BLEU score of $53.96$ 
($\approx+5$ BLEU points).
It seems plausible that the NMT system could be tuned to match the reference UN phrase-based system but this is beyond the scope of this work.

\subsection{Unsupervised performance}
The last two lines of Table~\ref{tab:results-3lang} concern the low-resource pair (es-fr)
evaluation which is the main focus of this work. 
NMT-0 performs worse than we expected between Spanish and
French: 10.02 es$\rightarrow$fr and 6.25 fr$\rightarrow$es. In
particular, we notice that NMT-0 often translates, partially, to English or the source language.
The zero-shot dual system (Dual-0) starts training from the parameters of the NMT-0 model.
We observe a marked improvement for the zero-shot translation directions, outperforming the baseline by $25.52$ and $32.58$ BLEU points for es$\rightarrow$fr and fr$\rightarrow$es respectively. By inspection, the translation quality seems quite
reasonable. Example translations of different models are discussed in the
Appendix \ref{ch:appendix}.
We also observe "catastrophic forgetting"~\citep{french1999catastrophic}: BLEU scores in the
supervised translation directions degrade, as no parallel data for
these pairs is seen during the incremental training. This happens
for all incrementally trained models.

The most important comparison is with respect to the fully supervised setting.
For this purpose,
we resume training the NMT-0 model
with the full set of $1$M es-fr parallel data. 
The resulting model is labeled NMT-F in Table \ref{tab:results-3lang}.
With respect to forgetting, we notice that, by never
decoding to English, the NMT-F performance degrades more in this
language. Interestingly, the dual model forgetting is more
uniformly distributed across all original language pairs.
However, NMT-F improves es$\rightarrow$fr 
from $10.02$ to $37.67$ and fr$\rightarrow$es, from $6.25$ to
$40.85$ BLEU points.

The results of the zero-shot dual model (Dual-0) are thus
close to two BLEU points away in both cases, 
without seeing any parallel sentence for the es-fr pair. 
To the best of our knowledge, this is the closest unsupervised machine
translation has come to the performance of the supervised setting. 
Although not directly comparable, this setup is similar to the Portuguese to Spanish evaluation in~\citep{johnson2016google}. Here, the best zero-shot model (model 2) gets a BLEU score of 24.75 
against 31.50 of the fully supervised NMT.

\citet{johnson2016google} showed that zero-shot translation can benefit from small amounts of parallel data. 
We simulate this case with the NMT-S model
which is trained additionally on 10k es-fr aligned sentences. The improvement is
considerable as NMT-S is only 2.52 (fr $\rightarrow$ es) and 4.57
BLEU points (es $\rightarrow$ fr) from the ``fully-supervised''
NMT-F. Nevertheless, NMT-S is still outperformed by the Dual-0 model.
The improvements carry over, to a smaller degree, also to the zero-shot dual model
(Dual-S) trained on the 10k es-fr sentences which gets its performance closer to fully supervised setting for both language pairs. 

\subsubsection{Extension to more languages}

\begin{table}
  \centering
  \begin{tabular}{lcccc}
    \toprule
                   & \bf Phrase-based             & \bf NMT-0                 & \bf NMT-F                 & \bf Dual-0 \\
                   \midrule
                   Aligned& en-fr, en-es, en-ru & en-fr, en-es, en-ru  & en-fr, en-es, en-ru  &  en-fr, en-es, en-ru           \\
                   &  en-fr, en-es, en-ru                        &                      & es-fr, es-ru, fr-ru  &              \\
                   & (11M each) & (1M each) & (1M each) & (1M each) \\
                   \midrule
                   Monol. & & & &es, fr, ru (0.5M each)\\
    \midrule
    en$\rightarrow$es & 61.26   & 47.51  & 44.96 \cmmnt{44.46} & 44.30 \\ 
    es$\rightarrow$en & 59.89   & 48.56  & 42.75 \cmmnt{38.95} & 45.55 \\ 
    en$\rightarrow$fr & 50.09   & 36.70  & 34.27 \cmmnt{34.53} & 34.34 \\ 
    fr$\rightarrow$en & 52.22   & 40.75  & 34.99 \cmmnt{30.97} & 37.75 \\ 
    en$\rightarrow$ru & 43.25   & 30.45  & 29.97 \cmmnt{29.30} & 29.47 \\ 
    ru$\rightarrow$en & 52.59   & 39.35  & 37.09 \cmmnt{34.04} & 37.96 \\ 
    \midrule
    \midrule
    es$\rightarrow$fr & 52.44   & 25.85  & 36.65 \cmmnt{36.50} & 34.51 \\ 
    fr$\rightarrow$es & 49.79   & 22.68  & 40.19 \cmmnt{39.85} & 37.71 \\ 
    es$\rightarrow$ru & 39.69   &  9.36  & 26.55 \cmmnt{26.46} & 24.55 \\ 
    ru$\rightarrow$es & 49.61   & 26.26  & 35.98 \cmmnt{35.88} & 33.23 \\ 
    fr$\rightarrow$ru & 36.48   &  9.35  & 24.50 \cmmnt{24.22} & 22.76 \\ 
    ru$\rightarrow$fr & 43.37   & 22.43  & 29.47 \cmmnt{29.27} & 26.49 \\ 
    \bottomrule
  \end{tabular}
    \caption{BLEU score results for the experiments with four languages.}
    \label{tab:results-4lang}
   
\end{table}

We experimented also with four languages, by adding Russian to the set. 
We follow the same procedure to sample the training data 
and build the shared vocabulary covering four languages.
We train the NMT baseline on $1$M parallel sentences from each pair: en-es, en-fr and en-ru. 
Thus, any translation direction that does not involve English is an
unsupervised translation direction.  
We train language models for all languages except English.
The purpose of this experiment is two-fold. 
On the one hand, we want to analyze how our approach works for languages with different alphabets. 
On the other hand, we want to see how it performs when there are multiple language pairs to be improved.


Table \ref{tab:results-4lang} summarizes the results. 
Regarding the supervised translation directions, NMT-0
maintains the relative performance compared to the phrase-based results, 
that is, it is still $10$-$13$ BLEU points lower, 
as in the previous experiment.
However, the zero-shot directions which do not have Russian as the target language 
obtain a good performance comparing to the experiment with three languages. 
We do not have yet a good explanation for this phenomenon. 
Interestingly, the NMT base model performs much better also with Russian as the source language. While translating to Russian performs poorly. 

Similar to the previous experiment, we increment the baseline NMT's training with $1$M parallel sentences from each pair es-fr, es-ru and fr-ru so as to have an estimate of the headroom with respect to the
supervised setting at comparable capacity and training schedule
(NMT-F in Table \ref{tab:results-4lang}).
The last column in Table \ref{tab:results-4lang} shows the results of the model after the zero-shot dual learning, trained on $1.5$M sentences from monolingual data, 
$500$K in each Spanish, French and Russian.
We can see that the performance is highly improved in this case as well. 
Moreover, it is consistent with what we saw in the previous experiment, 
i.e. the BLEU score that every zero-shot translation direction obtains is 
 $2$-$3$ BLEU points below of the supervised score of the base model. 

%% file: conclusion.tex

\section{Conclusion}
\label{ch:conclusion}
We propose an approach to zero-shot machine translation between language pairs for which there is no aligned data.
We build upon a multilingual NMT system \cite{johnson2016google} by applying reinforcement learning, using only monolingual data on the zero-shot translation pairs, inspired by dual learning~\cite{he2016dual}. 

Experiments show that this approach comes close to the performance of the corresponding supervised setting, for unsupervised language pairs.
Zero-shot dual learning outperforms the multilingual NMT baseline model even when a small parallel corpus for the zero-shot language pair is available. 
Finally, we have shown that our model can easily scale up to improve the zero-shot translation of multiple language pairs, 
yielding an improvement of up to $15.19$ BLEU points over the base NMT model, 
even when languages belong to a completely different family.

These results show that this is a promising framework for machine translation for low-resource languages, given that aligned data already exists for numerous other language pairs. This framework seems particularly promising in combination with techniques like bridging, given the abundant data, to and from English and other popular languages.
For future work, we would like to understand better the relation between
model capacity, number of languages and language families, to optimize 
information sharing. Exploiting more explicitly cross-language generalization; e.g., in  combination with recent unsupervised methods for embeddings
optimization~\citep{lample2017unsupervised,artetxe2017unsupervised}, seems also promising.
Finally, we would like to extend the reinforcement learning approach
to include other reward components; e.g., linguistically motivated.

%% file: qualitative.tex
\section{Qualitative Analysis}
\label{ch:appendix}

\subsection{Experiment with Three Languages}

Here we analyze the translation quality of the translations 
for the test set before and after the zero-shot dual learning algorithm.
Some examples are shown in Table \ref{tab:results-qualit3}. 
Examples are separated by a double horizontal line. 
For each, the source, the translation from the multilingual NMT baseline (trained on en-es and en-fr data), the translation after the zero-shot dual learning and the corresponding reference are provided.

After analyzing the output translations for both directions of the zero-shot pair on a high level,
we have noticed that the multilingual NMT baseline tends to translate to English, 
approximately half of the times. 
Apart from these, 
many sentences are rewritten in the source language.
Still, there is also a number of sentences that are correctly translated to the target language,
already with the multilingual NMT baseline model.
We have observed that these sentences are in general short or have a specific structure,
for example, meeting dates.

If we analyze the translations obtained after the reinforcement learning algorithm, instead,
we could see that all outputs are in the target language; 
just a few sentences contain some word that is not correctly translated. 
This means that our approach is capable of training the model to translate to the required language, 
even without any parallel data for the zero-shot language pair.

\begin{center}
\begin{longtable}{ll}  
  \caption{Qualitative analysis of the translations for the experiment with 3 languages. Examples are separated by a double horizontal line. For each, the first row shows the source sentence, the second is the outcome of the multilingual NMT baseline NMT-0 (trained on en-es, en-fr data), the third line shows the translation of the model after the zero-shot dual learning Dual-0 and the last row shows the reference.}
  \endfirsthead
  \multicolumn{2}{r}{{Continued on next page}} \\ \hline
  \endfoot
  \endlastfoot
    \midrule
    \textbf{(1) Source} & \textbf{L'EIIL a mis en ligne des vidéos dans lesquelles on peut voir des } \\ 
    & \textbf{personnes subir toute une série de châtiments abominables : (...) } \\
    NMT-0 & L'EIIL a mis en place des videos en indiquant que les personnes \\ 
    & séroviennent d'un ensemble d'abominables , (...)\\
    Dual-0 & La EIIL puso en marcha videos en los que se puede ver a \\
    & personas que padecen toda una serie de castigos abominables: (...) \\
    \textbf{Reference} & \textbf{El propio EIIL ha publicado vídeos en los que se muestran } \\ 
    & \textbf{imágenes de personas sometidas a una serie de castigos atroces, (...)} \\
    \midrule
    \midrule 
    \textbf{(2) Source} & \textbf{El propio EIIL ha publicado vídeos en los que se muestran imágenes} \\ 
    & \textbf{de personas sometidas a una serie de castigos atroces, (...)} \\
    NMT-0 & L'EIIL itself a publié des videos en indiquant des images de personnes \\ 
    & faisant l'objet d'une série de atroces châtiments , (...) \\
    Dual-0 & L'ensemble EIIL a publié des vidéos en indiquant des images de personnes \\
    & soumises à une série de châtiments atroces, (...) \\
    \textbf{Reference} & \textbf{L'EIIL a mis en ligne des vidéos dans lesquelles on peut voir des } \\ 
    & \textbf{personnes subir toute une série de châtiments abominables : (...)} \\
    \midrule
    \midrule
    \textbf{(3) Source} & \textbf{L'UNICEF a offert une aide en espèces d'urgence à des dizaines} \\ 
    & \textbf{ de milliers de familles dans les camps de déplacés (...).} \\
    NMT-0 & UNICEF provided emergency assistance to tens of thousands of \\ 
    & families in camps of internally displaced persons (...) . \\
    Dual-0 & El UNICEF ha ofrecido asistencia en efectivo de emergencia a decenas \\
    &  de miles de familias en los campamentos de desplazados internos (...) . \\
    \textbf{Reference} & \textbf{El UNICEF pagó ayudas monetarias de urgencia para asistir a } \\ 
    & \textbf{decenas de miles de familias desplazadas en los campamentos (...).} \\
    \midrule
    \midrule
    \textbf{(4) Source} & \textbf{Tenue de réunions périodiques par téléconférence pour coordonner } \\ 
    & \textbf{la planification conjointe des événements majeurs;} \\
    NMT-0 & Continuación de reuniones periódicas de videoconferencia para \\ 
    &  coordinar la celebración de seminarios de planificación de los \\ 
    & principales acontecimientos; \\
    Dual-0 & Celebración de reuniones periódicas por teleconferencia para coordinar \\
    & la planificación conjunta de los acontecimientos principales; \\
    \textbf{Reference} & \textbf{Reuniones periódicas por teleconferencia para coordinar  } \\ 
    & \textbf{la planificación conjunta de los eventos principales;} \\
    \midrule
    \midrule
    \textbf{(5) Source} & \textbf{Elles ont en outre aveuglément frappé des zones résidentielles, } \\ 
    & \textbf{y compris des camps de réfugiés, (...).} \\
    NMT-0 & En addition, a menudo se abusan de camas de zonas, \\
    & y compris les camps de réfugiés , (...). \\
    Dual-0 & También han armado ataques a las zonas residenciales, \\
    & incluidas campamentos de refugiados, (...). \\
    \textbf{Reference} & \textbf{También han atacado de manera indiscriminada zonas} \\
    &  \textbf{residenciales, incluidos los campamentos de refugiados, (...).} \\
    \midrule
    \midrule
    \textbf{(6) Source} & \textbf{Que se celebrará el jueves 2 de abril de 2015 a las 10.15 horas} \\
    NMT-0 & Que se tiendra le mardi 2 avril 2015 , à 10 h 15 \\
    Dual-0 & Qui se tiendra le jeudi 12 mai 2015, à 15 heures \\
    \textbf{Reference} & \textbf{Qui se tiendra le jeudi 2 avril 2015, à 10 h 15} \\
    \midrule
    \midrule
    \textbf{(7) Source} & \textbf{La supresión de la tasa de cambio mínima frente al euro llevó}  \\
    & \textbf{aparejado un nuevo movimiento negativo de las tasas de interés (...).} \\
    NMT-0 & The abolition of the minimum exchange rate compared to the euro \\
    & resulted in a new negative movement of interest rates (...). \\
    Dual-0 & La suppression de la croissance minimale face au euro a conduit à un \\
    & nouveau mouvement négatif des taux d'intérêt (...). \\
    \textbf{Reference} & \textbf{La suppression du taux plancher s'est accompagnée d'une baisse} \\
    & \textbf{du taux d'intérêt déjà négatif (...).} \\
    \midrule
  \label{tab:results-qualit3}
\end{longtable}
\end{center}

The first two examples from Table \ref{tab:results-qualit3} show the outcome of the two translation directions for the same sentence pair.
On the former case, 
NMT-0 rewrites the sentence in French, 
but the zero-shot dual model completely recovers the sentence to Spanish. 
On the second one, the base NMT model already translates the sentence fairly well to French, 
even though it contains some English word (\textit{itself, video}). 
Although the zero-shot Dual-0 avoids the English words, 
it does not get to improve much more.

Examples 3 and 7 show how the zero-shot dual learning obtains 
a translation outcome on the target language 
even when the base NMT-0's translation is in English, while Example 5 shows the same result for when the NMT-0 model's outcome is a mixture of languages.

The fourth example shows the improvement the Dual-0
obtains over the baseline NMT's translation in the target language, 
by removing terms that do not belong to the reference.

Finally, Example 6 provides a sample of the sentences that, 
in general, are correctly translated by the base NMT-0.
In this case, the day of the meeting is incorrect in the baseline's translation. 
Dual-0 fixes that along with the first word, 
but introduces a mistake on the month and on the time.

\subsection{Experiment with Four Languages}

We saw in the BLEU scores of the experiment involving four languages that the zero-shot translation of the baseline NMT is good for all directions that do not have Russian as target. 

For es$\rightarrow$ru and fr$\rightarrow$ru, we have noticed that, similar to what happened with the zero-shot translation on the previous experiment, almost half of the sentences are translated to English. This could explain the similar BLEU score that these two directions and the zero-shot translations from the previous experiment obtain.
However, the rest, a little more than half of the test set, is correctly decoded to Russian. Unlike in the previous experiment, there is no sentence that gets rewritten in the source language.

Holding to the behavior of the zero-shot translations in the previous examples, the translations that are correct are in general short sentences.
However, most of the Russian translations do not coincide with the corresponding references.
This explains the low BLEU score, even though half of the dataset is correctly decoded to Russian.

In contrast, regarding the translation directions that have Russian as the source language, most of the sentences are correctly decoded to the corresponding target language. There is a small number of sentences that get translated to English.
Interestingly, Russian to Spanish translation outputs contain some French sentences, while Russian to French barely contains any Spanish sentence, although it does have some Spanish words.

Finally, es$\leftrightarrow$fr translations are in general correctly translated to the target language. A small amount of sentences are translated into English and an even smaller amount gets rewritten in the source language.

The output of the zero-shot dual model holds the same as in the previous experiment; the target language dominates the translation outputs. English completely disappears from the translation outputs, except for some proper names that confuse the model.

\begin{center}
\begin{longtable}{ll}  
  \caption{Qualitative analysis of the translations for the experiment with 4 languages. Examples are separated by a double horizontal line. For each, the first row shows the source sentence, the second is the outcome of the multilingual GNMT baseline (trained on en-es, en-fr data), the third line shows the translation of the model after the zero-shot dual learning and the last row shows the reference.}
  \endfirsthead
  \multicolumn{2}{r}{{Continued on next page}} \\ \hline
  \endfoot
  \endlastfoot
    \midrule
    \textbf{(1) Source} & \textbf{Servicios de Gestión Estratégica}  \\
    NMT-0 & Strategic Management Services \\
    Dual-0 & Services de gestion stratégique \\
    \textbf{Reference} & \textbf{Services de gestion stratégique} \\
    \midrule
        \midrule
    \textbf{(2) Source} & \textbf{1. Aprobación del orden del día.}  \\
    NMT-0 & 1. Adoption du jour de l'ordre du jour. \\
    Dual-0 & 1. Adoption du jour. \\
    \textbf{Reference} & \textbf{1. Adoption de l'ordre du jour.} \\
    \midrule
        \midrule
    \textbf{(3) Source} & \textbf{Dans sa résolution adoptée en 2014 intitulée \guillemotleft Renforcement de l'efficacité et}  \\
    & \textbf{l'amélioration de l'efficience des garanties de l'Agence \guillemotright, la Conférence (...).} \\
    NMT-0 & Dans sa résolution adoptée en 2014 intitulée \guillemotleft Renforcement de l'efficacité et \\
    & l'amélioration de l'efficience des garanties de l'Agence \guillemotright, la Conférence (...). \\
   Dual-0 & En su resolución aprobada en 2014 titulado "Fortalecimiento de la eficiencia y \\
   & mejora de la eficiencia de las salvaguardias de la Autoridad", la Conferencia (...) \\
    \textbf{Reference} & \textbf{En su resolución de 2014 relativa al fortalecimiento de la eficacia y aumento } \\
    & \textbf{de la 
eficiencia de las salvaguardias del Organismo, la Conferencia (...).} \\
    \midrule
        \midrule
    \textbf{(4) Source} & \textbf{La Oficina se propone fomentar y mantener una cultura institucional de}  \\
    & \textbf{ética y rendición de cuentas, con el fin de aumentar tanto la credibilidad} \\
    & \textbf{como la eficacia de las Naciones Unidas.} \\
    NMT-0 & 31. \ru {Канцелярия планирует поощрять} and maintain a corporate culture and \\
    & accountability culture, with a view to increasing the credibility and  \\
    &  effectiveness of the United  Nations. \\
    Dual-0 & \ru {Управление предлагается поощрять и поддерживать институциональную} \\
    & \ru {культуру этики и подотчетности, с тем чтобы повысить доверие и} \\
    & \ru {эффективность Организации Объединенных Наций.} \\
    \textbf{Reference} & \textbf{\ru {Бюро призвано формировать и поддерживать корпоративную культуру}} \\
    & \textbf{\ru {строгого соблюдения этических норм и подотчетности с целью укрепления}} \\
    & \textbf{\ru {авторитета и эффективности Организации Объединенных Наций.}}
\\
    \midrule
    \midrule
    \textbf{(5) Source} & \textbf{\ru {3. Форум рекомендует Продовольственной и сельскохозяйственной}} \\
    & \textbf{\ru {организации Объединенных Наций (ФАО) в координации с коренными}} \\
    & \textbf{\ru {народами организовывать учебные курсы и другие мероприятия (...).}}
    \\
    NMT-0 & 1. Le Forum Permanent Forum on Indigenous Issues congratuce the \\
    & International Fund for Agricultural Development (IFAD) for its efforts in \\
    & the area of rural development to address the problems of food and hunger (...). \\
    Dual-0 & 3. El Foro recomienda a la Organización de las Naciones Unidas para la \\
    & Agricultura y la Alimentación (FAO) en coordinación con los pueblos \\
    & indígenas organizando cursos de capacitación y otras actividades (...). \\
    \textbf{Reference} & \textbf{El Foro recomienda que la Organización de las Naciones Unidas para la} \\
    & \textbf{Alimentación y la Agricultura (FAO), en coordinación con los pueblos} \\ 
    & \textbf{indígenas, organice cursos de formación y otras actividades (...).} \\
    \midrule
        \midrule
    \textbf{(6) Source} & \textbf{Les initiatives de la Chambre de commerce internationale ont contribué} \\ 
    & \textbf{indirectement à l'atteinte des OMD.}  \\
    NMT-0 & \ru {Инициативы} of the International Chamber of Commerce have contributed \\
    & indirectly to the detriment of the MDGs. \\
    Dual-0 & \ru{Инициативы Палаты международной торговли внесли существенный} \\
    & \ru{вклад в ущерб ЦРДТ.} \\
    \textbf{Reference} & \textbf{\ru {Инициативы Международной торговой палаты косвенно внесли}} \\ 
    & \textbf{\ru {вклад в достижение целей в области развития, сформулированных}} \\ 
    & \textbf{\ru {в Декларации тысячелетия.}} \\
    \midrule
    \midrule
    \textbf{(7) Source} & \textbf{Les activités de l'organisation sont principalement axées sur la} \\
    & \textbf{sensibilisation aux violations des droits de l'homme dans le monde. }  \\
    NMT-0 & \ru {Деятельность Организации, главным образом}, focuses on \\ 
    & raising awareness of human rights violations worldwide. \\
    Dual-0 & \ru {Деятельность организации главным образом уделяет повышению} \\
    & \ru {информированности о нарушениях прав человека во всем мире.} \\
    \textbf{Reference} & \textbf{\ru {Деятельность Института главным образом направлена на}} \\ 
    & \textbf{\ru {повышение осведомленности о нарушениях прав человека в}} \\ 
    & \textbf{\ru {разных странах мира.}} \\
    \midrule
    \midrule
    \textbf{(8) Source} & \textbf{\ru {Мы ожидаем в этой связи результаты глобального исследования}} \\
    & \textbf{\ru {в отношении детей, лишен
ных свободы;}}  \\
    NMT-0 & Nous attendons en conséquence les résultats de l'étude mondiale \\
    & concernant les enfants victimes de liberté ; \\
    Dual-0 & Nous attendons en conséquence les résultats de l'étude mondiale \\
    & concernant les enfants privés de liberté ;  \\
    \textbf{Reference} & \textbf{Nous attendons à cet égard avec intérêt les résultats de l'enquête} \\
    & \textbf{mondiale sur les enfants privés de liberté;} \\
    \midrule
  \label{tab:results-qualit4}
\end{longtable}
\end{center}

Table \ref{tab:results-qualit4} gives a few examples that allow us to evaluate all zero-shot translation directions. The first two examples show two scenarios for the es$\rightarrow$fr translation. The first one is translated to English by the multilingual NMT baseline model while the second one is translated already to French.

The third example shows how the zero-shot translation from NMT-0 basically copies the source sentence in the whole part that is shown. However, the Dual-0 model gets a good translation.

The directions that have Russian as target (see Examples 4, 6 and 7) show a mixture between English and Russian in the baseline NMT model. Still, our zero-shot dual model is capable of learning fairly good Russian translations for the given source sentences.

It is really interesting that the translation directions that have Russian as the source language (see Examples 5 and 8) never get rewritten in Russian by NMT-0; they mostly get translated to either English or the target language.

This analysis explains the difference in BLEU score that the multilingual NMT baseline and the zero-shot dual system obtain. Furthermore, it shows the effect that our approach has on the vanilla zero-shot translation, even when multiple zero-shot translation directions are involved, proving the potential of our approach.